\relax
\documentclass[letterpaper]{article} 
\usepackage{aaai22}  
\usepackage{times}  
\usepackage{helvet}  
\usepackage{courier}  
\usepackage[hyphens]{url}  
\usepackage{graphicx} 
\usepackage{natbib}  
\usepackage{caption} 
\DeclareCaptionStyle{ruled}{labelfont=normalfont,labelsep=colon,strut=off} 
\usepackage{algorithm}
\usepackage{algorithmic}
\usepackage{amsmath}
\usepackage{booktabs}

\usepackage{newfloat}
\usepackage{listings}
\floatstyle{ruled}
\newfloat{listing}{tb}{lst}{}
\floatname{listing}{Listing}
\title{eVAE: Evolutionary Variational Autoencoder}
\author {
    Zhangkai Wu,\textsuperscript{\rm 1}
    Longbing Cao, \textsuperscript{\rm 1}
    Lei Qi \textsuperscript{\rm 2}
}
\affiliations {
    \textsuperscript{\rm 1} University of Technology Sydney\\
    \textsuperscript{\rm 2} Southeast University\\
    berenwu1938@gmail.com, Longbing.Cao@uts.edu.au, qilei@seu.edu.cn
}

\usepackage{bibentry}

\begin{document}

\maketitle

\begin{abstract}
The surrogate loss of variational autoencoders (VAEs) poses various challenges to their training, inducing the imbalance between task fitting and representation inference. To avert this, the existing strategies for VAEs focus on adjusting the tradeoff by introducing hyperparameters, deriving a tighter bound under some mild assumptions, or decomposing the loss components per certain neural settings. VAEs still suffer from uncertain tradeoff learning.
We propose a novel \textit{evolutionary variational autoencoder} (eVAE) building on the variational information bottleneck (VIB) theory and integrative evolutionary neural learning. eVAE integrates a variational genetic algorithm into VAE with variational evolutionary operators including variational mutation, crossover, and evolution. Its inner-outer-joint training mechanism synergistically and dynamically generates and updates the uncertain tradeoff learning in the evidence lower bound (ELBO) without additional constraints. Apart from learning a lossy compression and representation of data under the VIB assumption, eVAE presents an evolutionary paradigm to tune critical factors of VAEs and deep neural networks and addresses the premature convergence and random search problem by integrating evolutionary optimization into deep learning. Experiments show that eVAE addresses the KL-vanishing problem 
for text generation with low reconstruction loss, generates all disentangled factors with sharp images, 
and improves the image generation quality,
respectively. eVAE achieves better reconstruction loss, disentanglement, and generation-inference balance than its competitors.
\end{abstract}

 \section{Introduction}
\label{sec:intro}

Variational autoencoders \cite{kingma2013auto} (VAEs) have attracted significant interest for their capability of learning continuous and smooth distributions from observations by integrating probabilistic and deep neural learning principles. However, VAEs still face some significant issues, including  dynamic uncertainty learning during variational inference and the tradeoff between representation compression and task fitting, despite various recent VAE mechanisms and variants. Below, we briefly analyze these issues and the gaps of existing solutions. A novel evolutionary VAE (eVAE) framework is then introduced to address some of these issues and gaps.

\subsubsection{VAEs and Gap Analysis}
\label{subsubsec:vae-gaps}

VAEs have demonstrated significant advantages of incorporating prior knowledge, mapping inputs to probabilistic representations, and approximating the likelihood of outputs. The integration of stochastic gradient variational Bayes (SGVB) estimator \cite{kingma2013auto} with neural settings learns a narrow probabilistic latent space to infer more representative attributes in the hidden space. VAEs have been applied in various domains, including time series forecasting \cite{fortuin2018som}, out-of-domain detection in images \cite{nalisnick2018deep,ran2022detecting,havtorn2021hierarchical,xiao2020likelihood}, generating images with spiking signals \cite{kamata2021fully}, and generating text by language modeling \cite{zhang2019improve}. 


Theoretically, the bound optimization of variational inference was introduced to substitute the log-likelihood function with a surrogate function to make it optimized by gradient descent. In practice, the evidence lower bound (ELBO) cannot approach the maximum of conditional likelihood with a close gap between posterior and prior. It causes dynamic uncertainty during inference and failed tradeoff between representation robustness and reconstruction effectiveness. Weak KL divergence could incur KL vanishing while strong KL divergence could result in bad likelihood. The tradeoff is sensitive to posterior distribution disjoint, data characteristics, and network architectures  \cite{bozkurt2019rate}. 


Recently, various techniques have been proposed to address these issues. The first set aims to adjust the balance in objective functions. Examples are $\beta$-VAE incorporating a hyperparameter $\beta$ \cite{higgins2016beta}, InfoVAE adding a scaling parameter to the KL term \cite{zhao2019infovae}, SA-VAE having a cyclical annealing schedule to progressively increase $\beta$ for reducing KL vanishing \cite{fu2019cyclical}, and ControlVAE introducing the proportional-integral-derivative (PID) control to tune the hyperparameter \cite{shao2020controlvae}. They are partial solutions only adjusting one part of the objectives, failing to weigh and resolve the balance issue in dynamic settings. Another set is to tune network architectures or VAE settings, e.g., VQ-VAE and their variants \cite{razavi2019generating,berliner2022learning}, spiking VAE \cite{kamata2021fully}, multimodal VAE  \cite{sutter2021generalized}, mixture-of-experts multimodal VAE \cite{shi2019variational} and tuning of their ELBO \cite{kviman2022multiple,jankowiak2021surrogate} building on the product of experts structure. However, these models are subject to specific tasks or data settings.

\textbf{eVAE} 
From the information bottleneck perspective, VAEs can be treated as a lossy information compression process. Adjusting the KL  within a range controls the information bottleneck flowing from representing latent variables to reconstructing samples, thus benefiting the trade-off between compression and reconstruction \cite{alemi2016deep,burgess2018understanding}. 
This variational information bottleneck inspires us to improve the VAE mechanism and objective function. 
To directly address the generation-inference balance in an evolving dynamic setting without constraints on networks and inputs, we introduce a novel evolutionary VAE (eVAE). eVAE incorporates and integrates variational evolutionary learning and variational information bottleneck into VAE to achieve better optimum exploration and the tradeoff between representation compression and generation fitting. 
However, integrating evolutionary learning into VAE is an open topic. We  propose the variational genetic algorithm into eVAE to optimize VAE objectives and its modeling exploration in an evolving manner.

Consequently, eVAE dynamically optimizes the VAE inference (the KL terms) in an evolutionary and probabilistic manner, 
which further tunes the VAE generation toward balancing generation and inference. To avert premature convergence, eVAE introduces probabilistic chromosome selection for smooth search. To avoid exhaustive random search, we apply the simulated binary crossover \cite{deb2001self} and Cauchy distributional mutation to guide the training toward a stable convergence. 
The main contributions of this work are as follows:
\begin{itemize}
    \item We propose the eVAE model to balance the tradeoff between generation and inference. Specifically, a Cauchy-based variational mutation operator and an SBX-based variational crossover operator are proposed to train a model with an evolving inner-outer-joint training algorithm, tackling the premature convergence and random search problem when optimizing deep models assisted with a variational genetic algorithm.
    \item We illustrate eVAE by deriving the lower bound on $\beta$-VAE related models under the information bottleneck theory. We analyze the information flow in the corresponding lower bound per the rate-distortion theory to show that only the iteration-specific lower bound in eVAE can capture the optimization trend to train effectively.
    \item We evaluate eVAE on three tasks: disentangled representation learning on dSprites, image generation on CelebA, and text generation on PTB corpus, respectively. 
\end{itemize}

eVAE forms the first evolutionary VAE framework without empowering constraints on VAE architectures, input settings, and objective function (ELBO). It integrates variational encoding and decoding, information bottleneck, and evolutionary learning into the deep neural framework. 
The evaluation against the state-of-the-art VAEs shows that eVAE achieves better disentanglement, reconstruction loss, and solve the KL vanishing effectively.

\section{VAE Background and Issues}
\label{sec:background}

Here, we briefly introduce the background of VAEs by focusing on their objectives and approaches to addressing the inference-generation balance since they are mostly relevant to this work. Other developments on VAEs such as adjusting VAE network structures, input settings, and learning constraints are excluded due to their irrelevance to this work.

\subsubsection{VAE}
\label{subsubsec:vae}

VAEs mitigate autoencoder issues like sparse representation \cite{tolstikhin2017wasserstein} by learning continuous and smooth representation distribution $p(x)$, $x \in \mathcal{X}$ from observations $\mathcal{X}$ over latent variables $\boldsymbol{z}$. After learning an encoding distribution $q_{\phi}(\boldsymbol{z} \mid \boldsymbol{x})$ in encoding neural networks, VAEs apply variational inference to approximate the posterior distribution $p_{\theta}(\boldsymbol{x} \mid \boldsymbol{z})$. Learning tasks such as reconstructed and generated outputs can then be sampled from this learned distribution in a generative process. With the SGVB estimator and reparameterization trick, the gradients become tractable, and the generative parameters $\theta$ and inference parameters $\phi$ are learnable. The objectives of VAEs can be converted to ELBO with the expectation over empirical distribution $p_{\text {data}}$ of the data towards both reconstruction $\mathcal{L}_{\mathrm{R}}$ and inference $\mathcal{L}_{\mathrm{I}}$ \cite{esmaeili2019structured,ghosh2019variational}.
\begin{equation}
\begin{aligned}\label{eq:vae}
\mathcal{L}_{E L B O} &=E_{x \sim p_{\text {data }}}\left[E_{q_{\phi}(\boldsymbol{z} \mid \boldsymbol{x})}\left[\log p_{\theta}(\boldsymbol{x} \mid \boldsymbol{z})\right]\right.\\
&\left.-\mathrm{KL}\left(q_{\phi}(\boldsymbol{z} \mid \boldsymbol{x}) \| p(\boldsymbol{z})\right)\right] \\
&=E_{x \sim p_{\text {data }}} \mathcal{L}_{\mathrm{R}}+\mathcal{L}_{\mathrm{I}}
\end{aligned}
\end{equation}

\subsubsection{VAE Issues}
\label{subsubsec:vae-mitigate}
To solve ELBO, the inference model $q_{\phi}(z|x)$ can be trained jointly by maximizing the ELBO to acquire reasonable compression for task fitting. However, a weak capacity of the decoder and the variety of data could make the expressive posterior favor task fitting rather than optimal inference \cite{zhao2019infovae}. For example, in variational language generation, the decoder built on autoregressive models such as LSTM and PixelCNN can generate language samples by the autoregressive property rather than the posterior-based latent variables \cite{wu2019solving}. The VAE degenerates to an autoregressive model where the KL divergence between posterior and prior reaches zero quickly during training. This results in KL vanishing and poor generalization in test for the lack of diversity. The approaches of learning orthogonal transformation of priors with the same distribution by the decoder \cite{khemakhem2020variational,mita2021identifiable} may sacrifice accurate inference in optimal representation and generalization of fitting the data. 
Other research attributes the training conflict to the inherent property of bound optimization. For example, under a solid factorial assumption about the posterior distribution \cite{locatello2019fairness}, i.e.,
\begin{equation}\label{eq:jointprob}
    p(\mathbf{x}, \mathbf{z})=p(\mathbf{x} \mid \mathbf{z}) \prod_{i} p\left(z_{i}\right),
\end{equation}
the ELBO constraining the variational samples favors the data fitting \cite{burda2015importance} but fails to maximize the probability mass on log-likelihood. In addition, the vanilla VAE optimizer strengthens the disjointness between $q_{\phi}(z|x_{i})$, i.e., $\mu_{i} \rightarrow \infty, \sigma_{i} \rightarrow 0^{+}$, to separate the log-likelihood concentrated on each sample, resulting in maximizing the mass of joint distribution \cite{zhao2019infovae}. 

\subsubsection{VAE Enhancement}
\label{subsubsec:vae-enhance}

One way to address the above issues is to tighten the log-likelihood lower bound for correct variational approximation in posterior \cite{alemi2018fixing,wang2019neural,domke2019divide,jankowiak2021surrogate,ruiz2021unbiased,kviman2022multiple}, prior \cite{tomczak2018vae,klushyn2019learning} and decomposition of ELBO \cite{zhao2019infovae,esmaeili2019structured} under some mild assumptions. For example, $\beta$-VAE \cite{higgins2016beta} adds the hyper-parameter $\beta$ to weigh the $\mathcal{L}_{KL}$ term. Then, ELBO minimizes $\mathcal{L}_{R}$ to the convergence of data fitting collectively with the regularization $\mathcal{L}_{I}$ by varying $\beta$:
\begin{equation}
\begin{aligned}\label{eq:beta-vae}
\mathcal{L}_{E L B O}=& E_{x \sim p_{d a t a}}\left[E_{q_{\phi}(\mathbf{z} \mid \mathbf{x})}\left[\log p_{\theta}(\mathbf{x} \mid \mathbf{z})\right]\right.\\
&\left.-\beta D_{K L}\left(q_{\phi}(\mathbf{z} \mid \mathbf{x}) \| p(\mathbf{z})\right)\right] \\
=& E_{x \sim p_{\text {data }}}\left[\mathcal{L}_{\mathrm{R}}+\beta \mathcal{L}_{\mathrm{I}}\right]
\end{aligned}
\end{equation}

$\beta$-VAE introduces some fundamental limitations, which trigger various follow-up research. InfoVAE introduces a scaling parameter $\lambda$ on the KL-term and converts the objective to \cite{zhao2019infovae}:
\begin{equation}
\begin{aligned}\label{eq:info-vae}
\mathcal{L}_{E L B O}=& \alpha I_{q}(\mathbf{x} ; \mathbf{z})-D_{K L}\left(q_{\phi}(\mathbf{z} \mid \mathbf{x}) \| p(\mathbf{z})\right) \\
&-E_{q(z)}\left[D_{K L}\left(q_{\phi}(\mathbf{z} \mid \mathbf{x}) \| p_{\theta}(\mathbf{x} \| \mathbf{z})\right)\right]
\end{aligned}
\end{equation}
where $I_q$ is the mutual information with weight $\alpha$ and $\alpha + \lambda - 1=0$. This linear tuning on the KL shows limitation to dynamic uncertainty.

Further, conditional VAE (CVAE) \cite{CVAE} introduces an initial guess as a conditional variable into the objective function for multimodal data. The SA-VAE involves a cyclical annealing schedule to split the training to multiple cycles starting at $\beta=0$ and progressively increases $\beta$ until $\beta=1$ to reduce the KL vanishing \cite{fu2019cyclical}. In ControlVAE, the PID control compares the KL divergence with a set point, with their difference as feedback to the controller to tune the hyperparameter $\beta(t)$ \cite{shao2020controlvae}. ControlVAE thus optimizes KL dynamically but is constrained by the PID controller which follows a separate tuning mechanism from the VAE itself.

The existing work leaves gaps for building an approximate weight allocation between reconstruction and inference, tuning external hypberparameters within the VAE working mechanism and handling these issues in a dynamic manner over an evolutionary learning process. 
Our eVAE addresses these gaps by incorporating the variational genetic learning into balancing inference and generation and evolutionarily involving their effect into adjusting the VAE learning behaviors toward better uncertain tradeoff learning.

\begin{figure*}[htbp]
    \centering
    \includegraphics[width=1.6\columnwidth]{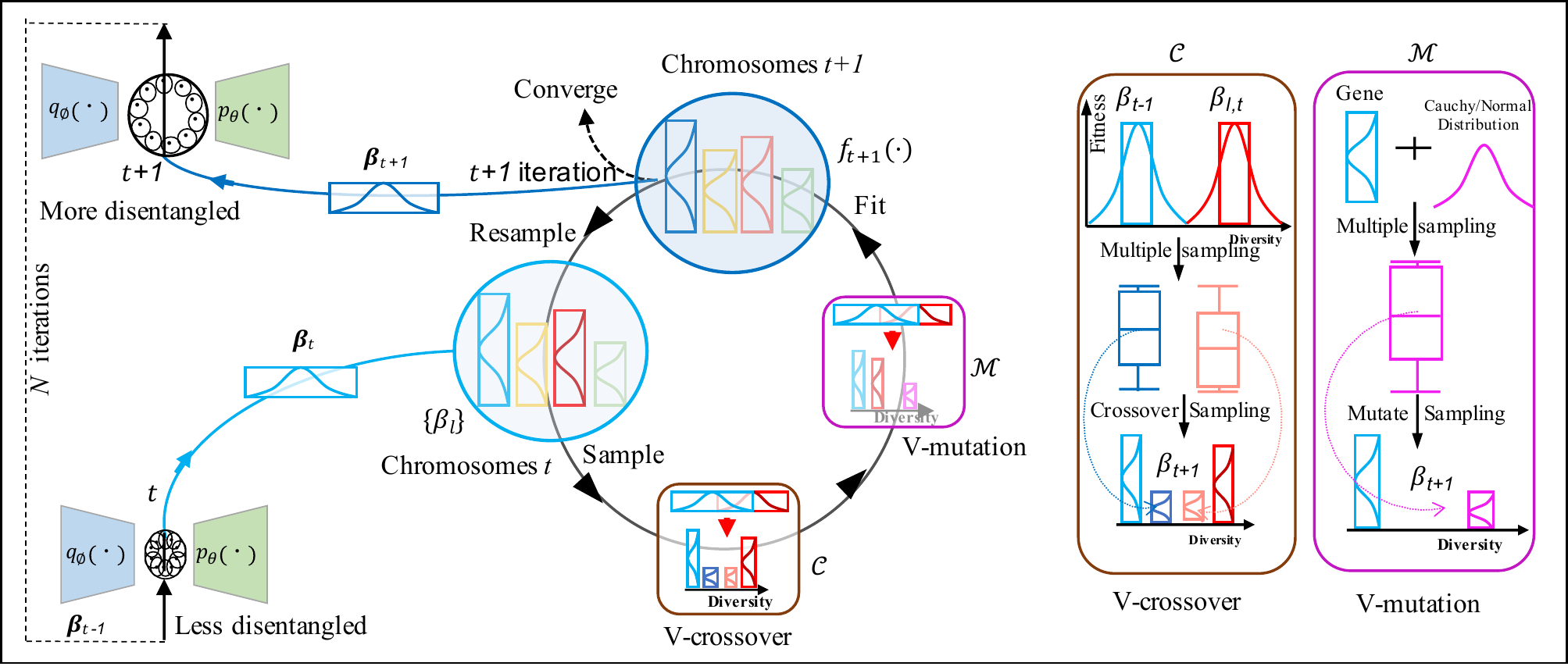}
    \caption{The framework of eVAE. The VAE results inform the chromosome sampling. The genes are then updated by variational V-crossover and V-mutation. The evolved results are checked per fitness for $t+1$ retraining, giving up, or converging.}
    \label{fig:eVAE-framework}
\end{figure*}

\section{The eVAE Model}

eVAE also tunes the representation-generation balance in Eq. (\ref{eq:vae}) by jointly addressing the following issues in Eqs. (\ref{eq:beta-vae}) and (\ref{eq:info-vae}): tuning hyperparameter in an outer circle irrelevant to VAE behaviors, non-dynamic optimization, and constrained settings on data and networks. 
Figure \ref{fig:eVAE-framework} illustrates the eVAE framework of variational evolutionary learning to improve the VAE balance.

\subsection{eVAE - Evolving Inner-outer-joint Training}
\label{subsec:evae-structure}

eVAE implements an evolving learning process of optimizing both VAE and its evolutionary parameters through an inner-outer-joint iterative training process, as shown in Figure \ref{fig:eVAE-infoflow}. At time $t$, given the input $x_t \in \mathcal{X}$, a VAE inner training process trains a VAE model $\mathcal{V}$ (we do not constrain the VAE framework, for generality, we use $\beta$-VAE for case study in this paper) with some initialization $\beta_0$ to learn the encoder $q_{\phi}(\mathbf{z}_t \mid \mathbf{x}_t)$, the posterior $p_{\theta}(\mathbf{x}_t \mid \mathbf{z}_t)$ with prior $p(\mathbf{z})$. We obtain the VAE inner objective function with initial weight $\beta_0$:
\begin{equation}
\begin{array}{r}\label{eq:e-vae-1}
\mathcal{L}_{E L B O_{t}}=E_{x_{t} \sim p_{\mathcal{X}}}\left[E_{q_{\phi}\left(\mathbf{z}_{t} \mid \mathbf{x}_{t}\right)}\left[\log p_{\theta}\left(\mathbf{x}_{t} \mid \mathbf{z}_{t}\right)\right]-\right. \\
\left.\beta_{0} D_{K L}\left(q_{\phi}\left(\mathbf{z}_{t} \mid \mathbf{x}_{t}\right) \| p(\mathbf{z})\right)\right]
\end{array}
\end{equation}

By applying SGVB and reparameterization trick, the generative process of $\mathcal{V}$ estimates the posterior as
\begin{equation}\label{eq:post}
    p_{\theta}(\mathbf{x} \mid \mathbf{z}) = \prod_t p_{\theta}(\mathbf{x}_t \mid \mathbf{z}_t)
\end{equation}
where $\mathbf{x}_t \mid \mathbf{z}_t \sim \mathcal{N}(\mu (\mathbf{z}_t; \theta), \Sigma(\mathbf{z}_t; \theta))$ with $\mathbf{z}_t = \mu_{x_t} + \mathbf{\epsilon} \Sigma_{x_t}$ and $\mathbf{\epsilon} \sim \mathcal{N}(0,I)$. 

Then, we introduce an outer variational evolutionary learner $\mathcal{E}$ (illustrated by a variational genetic algorithm introduced below on Variational Evolution) to optimize the objective in Eq. (\ref{eq:e-vae-1}) in an outer process. $\mathcal{E}$ with the fitness function $f$ samples a chromosome $\beta_t$ from an evolving distribution $\beta_t \sim \mathcal{R}$ and evolves it by taking variational crossover and mutation  to generate and improve parameters $\beta_{t+1}$. 

The evolved parameter is then incorporated into $\mathcal{V}_t$ for the next iteration $t+1$ training. With the relevant ELBO information back-propagated to tune the VAE network and optimize the relevant parameters, we obtain the updated reconstruction and inference performance. Consequently, Eq. (\ref{eq:e-vae-1}) is evolved to:
\begin{equation}
\begin{array}{r}\label{eq:evae-ga}
\mathcal{L}_{E L B O_{t+1}}=E_{x_{t+1} \sim p_{\mathcal{X}}}\left[E _ { q _ { \phi } ( \mathbf { z } _ { t + 1 } | \mathbf { x } _ { t + 1 } ) } \left[\operatorname { l o g } p _ { \theta } \left(\mathbf{x}_{t+1}\right.\right.\right. \\
\left.\left.\left.\mid \mathbf{z}_{t+1}\right)\right]-\beta_{t+1} D_{K L}\left(q_{\phi}\left(\mathbf{z}_{t+1} \mid \mathbf{x}_{t+1}\right) \| p(\mathbf{z})\right)\right]
\end{array}
\end{equation}

If the fitness Eq. (\ref{eq:evae-fitness}) is satisfied, we update $\beta$ by $\beta_{t+1}$. This then results in updated $\mathcal{V}_{t+1}$ and its parameters $\phi_{t+1}$, $\theta_{t+1}$ and $\beta_{t+1}$ for iteration $t+1$. We further repeat the above VGA $\mathcal{E}$ to obtain another set of parameters and retrain model $\mathcal{V}_{t+1}$ until it converges. 

The above inner-outer iterative training progresses over time and samples in the generative process to iteratively optimize the balance between reconstruction (i.e., minimizing $\mathcal{L}_{R} = \| x_t - \hat{x}_t \|$ with the reconstructed $\hat{x}_t$) and inference (i.e., seeking appropriate $\mathcal{L}_{I}$) and the VAE learning behaviors. The VAE $\mathcal{V}$ is then optimized until converges. 

Accordingly, the eVAE generative process optimizes the following objective function:
\begin{equation}
\begin{array}{r}
\mathcal{L}_{e V A E}=\min _{\boldsymbol{\theta}_{t}, \boldsymbol{\phi}_{t}} \sum_{t=1}^{N} \mathcal{L}_{E L B O_{t+1}}\left(\boldsymbol{\theta}_{t}, \boldsymbol{\phi}_{t}\right.; \\
\left.f\left(\mathcal{E}\left(\phi_{t}, \theta_{t}, \{\beta\}, \mathcal{L}_{E L B O_{t}}\right)\right), \mathbf{x}_{t}\right)
\end{array}
\end{equation}
to approach $\beta^{*}$ for the optimal balance between representation and construction: 
\begin{equation}
\beta^{*} \sim f\left(\mathcal{E}\left(\{\beta\} \mid \phi_{t}, \theta_{t}, \mathcal{L}_{E L B O_{t}}\right)\right)
\end{equation}

The overall ELBO of eVAE can then be represented by:
\begin{equation}
\begin{array}{r}\label{evae:loss}
\mathcal{L}_{e V A E}=E_{\mathbf{z} \sim q_{\phi}\left(\mathbf{z}_{t} \mid \mathbf{x}_{t}\right)} \log p_{\theta}(\mathbf{x}_{t} \mid \mathbf{z}_{t})+ \\
f_{\beta_{t} \sim R}(\beta_{t}, \mathcal{E}) D_{K L}(q_{\phi}\left(\mathbf{z}_{t} \mid \mathbf{x}_{t}\right) \| p(\mathbf{z})) \\
=E_{x_{t+1} \sim p_{\mathcal{X}}}[\mathcal{L}_{R}+f(\beta) \mathcal{L}_{I}+ \mathcal{E}(\Delta \mathcal{L}_{E L B O})] 
\end{array}
\end{equation}

\subsection{Variational Evolution in eVAE}
\label{subsec:var-evo}
Here, we introduce the variational evolutionary learner $\mathcal{E}$ optimizing the parameters. We instantiate $\mathcal{E}$ by a variational genetic algorithm (VGA) with its variational crossover and mutation operations. This avoids typical issues including premature convergence and random search during discordant exploitation and exploration.

\begin{figure}[!htp]
    \centering
    \includegraphics[width=0.89\columnwidth]{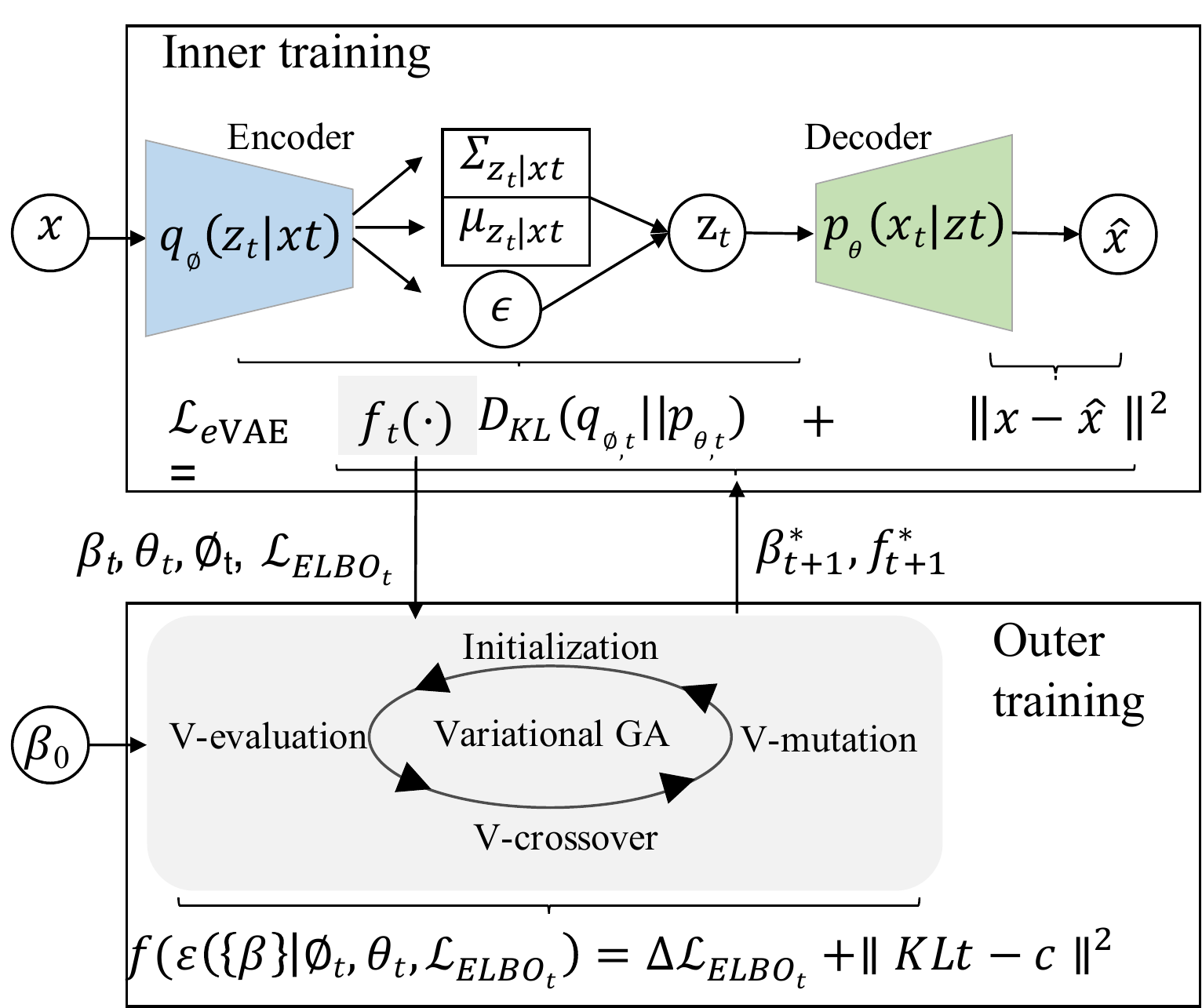}
    \caption{eVAE - inner-outer joint evolutionary training process. The upper part shows the VAE training at time $t$, the objectives are then incorporated into the lower part - the outer training by variational genetic algorithm, whose fitness-based optimized results are fed to the VAE for further training.}
    \label{fig:eVAE-infoflow}
\end{figure}

\subsubsection{Variational Genetic Algorithm}
\label{subsubsec:vga}

eVAE integrates the VGA-based outer parameter optimization combined with the VAE-internal gradient descent-based optimization. 
Accordingly, VGA consists of a few variational steps and operations, initialization, variational crossover (V-crossover), variational mutation (V-mutation), and variational evaluation (V-evaluation). As shown in the bottom part of Figure \ref{fig:eVAE-infoflow}, below, we introduce them respectively.


\textit{Chromosome selection}: 
To enable a larger and more smooth GA search space compatible with VAEs, chromosomes are embedded by a continuous variable $\beta$, sampled from an evolving distribution $\mathcal{R}$ (e.g., in the crossover). Assume $L$ individual chromosomes forming a candidate group $\{\beta_{l}\} = \{\beta_{1}, \dots, \beta_{L}\}$, a chromosome is chosen from this candidate group to train the VAE, which is associated with a fitness value $f$ after the VGA operations and VAE retraining. We thus have the chromosome-fitness pairs embedded for the VGA evolution and offspring selection over generations, such as:
\begin{equation}
     \{\beta,f\} =((\beta_{1},f_{1}), \dots, (\beta_{L},f_{L}))
\end{equation}
for optimizing the inner-outer joint training.

\textit{V-crossover}: Paired chromosomes at time $t-1$ and $t$ are crossovered to generate new genes for improving the genetic variety of offsprings. Following the simulated binary crossover (SBX) \cite{deb2001self} method, we implement the following crossover operations $\mathcal{C}$ to induce a large search space evolving over chromosome $\beta_{t-1}$ at time $t-1$ and chromosome $\beta_{t}$ selected from the chromosome candidate group $\{\beta_l\}$ at time $t$ to obtain the offspring candidate chromosome $\beta_{t+1}$ for the next time $t+1$ in $\mathcal{C}$ strategy:
\begin{equation}
  \mathcal{C}: \left\{\begin{array}{l}
  \beta_{t+1} = \frac{1}{2}[(1+r_{c})\beta_{l,t} + (1-r_{c})\beta_{t-1}], or \\
  \beta_{t+1} = \frac{1}{2}[(1-r_{c})\beta_{l,t} + (1+r_{c})\beta_{t-1} ]
  \end{array}\right.
\label{con:C1}
\end{equation}
where $r_{c}$ is the crossover rate, which is drawn from a probability density function $P_{c}(r_{c})$:
\begin{equation} \label{con:C2}
P_{c}\left(r_{c}\right)= \begin{cases}0.5(\eta+1) r_{c}^{n}, & \text { if } r_{c} \leq 1 \\ 0.5(\eta+1) \frac{1}{r_{c}^{\eta+2}}, & \text { otherwise }\end{cases}
\end{equation}
by concentrating their corresponding parents $\beta_{t-1}$ and $\beta_{l,t}$. By sampling $r_{c}$ from the following function: 
\begin{equation}\label{con:C3}
r_{c}= \begin{cases}(2 u)^{\frac{1}{\eta+1}}, & \text { if } u \leq 0.5 \\ \left(\frac{1}{2(1-u)}\right)^{\frac{1}{\eta+1}}, & \text { otherwise }\end{cases}
\end{equation}
where $u$ is a random variable, and $\eta$ is a hyper-parameter for scaling, the offsprings $\beta_{t+1}$ can then be created by selecting the better $\mathcal{C}$ strategy (i.e., better satisfying Eq. \eqref{eq:evae-fitness}).

\textit{V-mutation}: The crossover-generated offsprings can be further mutated to improve their genomic diversity and evolutionary capacity. Aligning with the V-crossover, a variational mutation strategy $\mathcal{M}$ is taken to diversify the offspring to update $\beta_{t+1}$ inherited from crossover or $\beta_{t}$ from the current moment. With a relatively small probability $p_{m}$, a chromosome $\beta_{t+1}$ selected from group is mutated as:  
\begin{equation}\label{con:M1}
    \mathcal{M}: \beta_{t+1} = \beta_{t+1} + r_{m}
\end{equation}
where $r_{m}$ is a random variable sampled from the Cauchy distribution $p_{m}$:
\begin{equation}\label{con:M2}
    P_{m} = \frac{1}{\pi}(\frac{1}{1 + \beta_{t+1}^{2}})
\end{equation}

\subsubsection{V-evaluation \& VGA fitness function}
\label{subsubsec:v-eva-fitness}

The chromosomes updated by the V-crossover and V-mutation operations are evaluated in alignment with the VAE objectives to determine whether they are transferable to the next generation. Within the VAE framework, we take the following heuristic fitness function $f$ to guide the evolution of chromesomes and their fitness to VAE objectives. The fitness function integrates the direction of VAE-oriented stochastic gradient descent and the distance to the optimal KL divergence to achieve multi-objectives.
\begin{equation}\label{eq:evae-fitness}
    f_{t+1}  = \Delta \mathcal{L}_{ELBO_{t+1}} + ||KL_{t+1}(\beta_{t+1}) - c||
\end{equation}
where $\Delta \mathcal{L}_{ELBO_{t+1}}$ is:
\begin{equation}\label{con:F2}
    \Delta \mathcal{L}_{ELBO_{t+1}} = \mathcal{L}_{ELBO_{t+1}}(\beta_{t+1}) - \mathcal{L}_{ELBO_{t}}(\beta_{t})
\end{equation}
denoting the VAE's ELBO difference after applying the evolved $\beta$. $c$ is the task-specific information bottleneck, which ensures the bound optimization of eVAE over the evolutionary optimization.

This V-evaluation and the fitness $f$ guide  eVAE to converge and balance the reconstruction and inference with enhanced evolutionary parameterization. The larger the fitness value, the stronger the chromosome in its candidate group, which forms a strong pair $\{\beta^{*},f^{*}\}$. 

The whole training processes of the eVAE model is illustrated in Algorithm \ref{algorithm:eVAE} in the Supplementary Material A.

\section{Theoretical Analysis}
\label{var-infobottle}

In this paper, eVAE takes $\beta$-VAE as a case study of evolutionary VAE. Hence, we analyze the effects of eVAE in adjusting $\beta$-VAE parameters, tradeoff between reconstruction and regularization, and training performance below.

$\beta$-VAE directly tunes hyperparameter $\beta$ setting a task-relevant ratio between the inference loss $\mathcal{L}_{\mathrm{I}}$ and the generation loss $\mathcal{L}_{\mathrm{R}}$. 
In essence, incrementally regularizing $\mathcal{L}_{\mathrm{I}}$ by $\beta$ is equivalent to maximizing the likelihood on an  experiment-specific threshold, i.e., a constant  $C$, \cite{higgins2016beta}:
\begin{equation}
\begin{aligned}\label{eq: theory_elbo}
\max _{\phi, \theta} E_{x \sim p_{\text {data }}}\mathcal{L}_{R}  \quad \text { s.t. } \beta \mathcal{L}_{I}<C
\end{aligned}
\end{equation}
This learning process can be interpreted in information bottleneck (IB) theory \cite{alemi2018fixing}. The $\beta \mathcal{L}_{I}$ constraint operates as the bottleneck, compressing the capacity of latent variable $Z$ organized from input $X$ to expressively represent the target $\hat{X}$ \cite{alemi2016deep}. Accordingly, IB maximizes $I(Z, \hat{X}) $ to acquire a concise representation:
\begin{equation} \label{eq: theory_IB}
\max I(Z, \hat{X}) \quad \text { s.t. } I(X, Z) \leq I_{c}
\end{equation}
Further, the variational information bottleneck (VIB) introduces an experiment-specific lower bound for optimization:
\begin{equation}\label{eq: theory_beta}
    \mathcal{L}_{\beta-VAE}(\theta, \phi)=-D-\beta_{C} R \leq -\beta_{C} I -D
\end{equation}
where $R$ refers to the compression rate, and $D$ refers to the distortion measuring the representation relevance to a task, and $\beta_{C}$ is an experiment-specific constant. VAEs can thus be viewed as a process of achieving the rate-distortion tradeoff.

ControlVAE is a variant of $\beta$-VAE, tuning the KL-specific $\beta_{KL}$ with a PID control to  a desired KL point:
\begin{equation}\label{eq: theory_control}
\mathcal{L}_{ControlVAE}(\theta, \phi)=-D-\beta_{KL} R \leq -\beta_{KL}I - D
\end{equation} 

eVAE extends $\beta$-VAE tuning the information bottleneck $\beta I(X,Z)$ by a variational evolutionary learner $\mathcal{E}$. After iteration $t$ of outer training, $\beta_{t}$ evolves to fit the task and training phase, guided by the fitness function $f$. eVAE  generates a lower bound to optimize  iteration $t+1$ of the inner training phase:
\begin{equation}
\begin{aligned}\label{eq: theory_evae}
\mathcal{L}_{e V A E}  
&\leq  -\mathcal{E}(\beta_{t})I_{t} -\mathcal{E}(\beta_{t}) D_{t}
\end{aligned}
\end{equation}
where an evolving iteration-specific lower bound can strike a compromise between task fitting in $-\mathcal{E}(\beta_{t}) D_{t}$ and inference quality in $\mathcal{E}(\beta_{t})I_{t}$.
The derivations of Eq. \eqref{eq: theory_beta} and Eq. \eqref{eq: theory_evae} can be found in the Supplementary Materials B and C.

\begin{figure}[!htp]
    \centering
    \includegraphics[width=0.75\columnwidth]{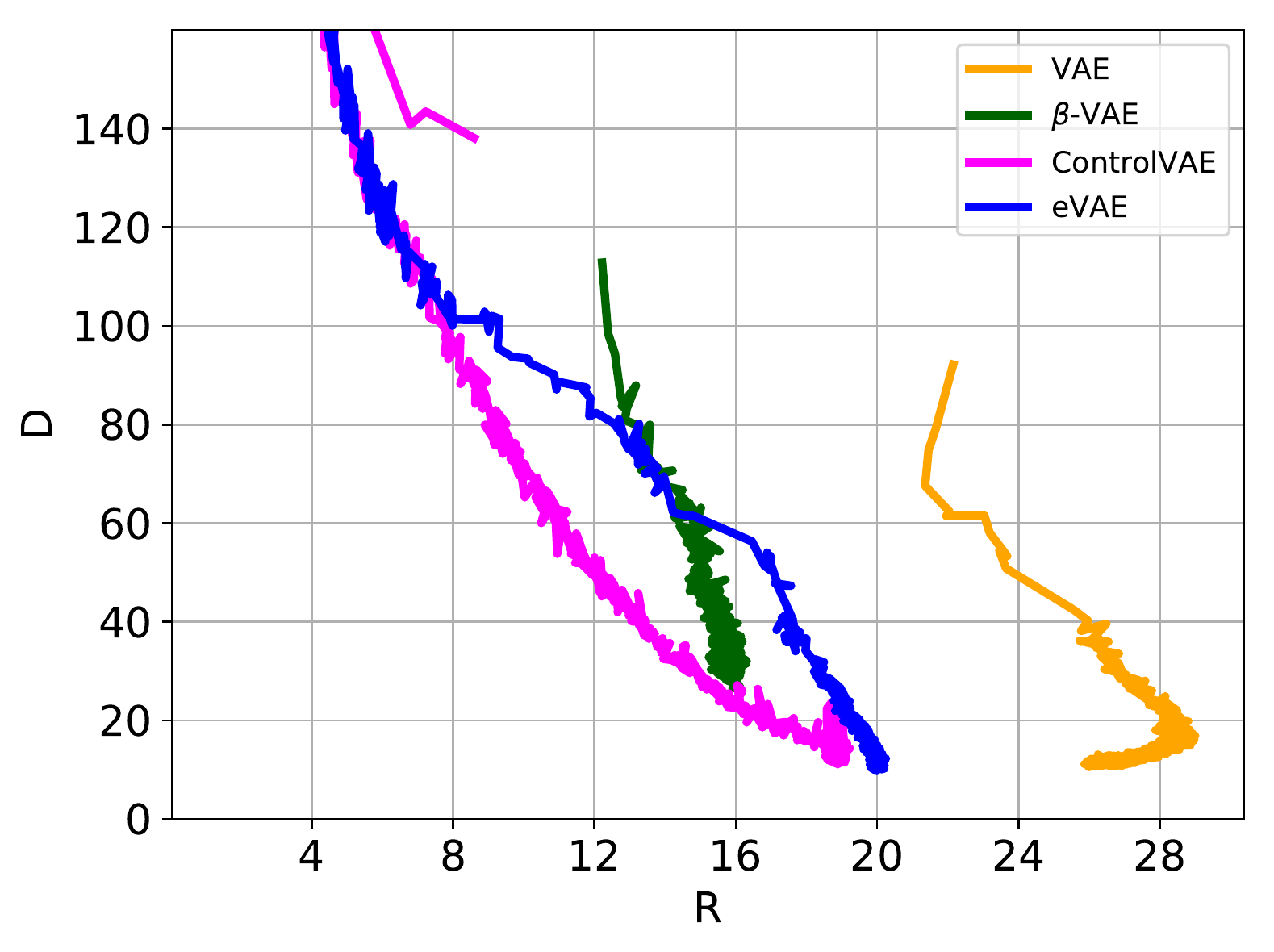}
    \caption{The information plane with the $R-D$ curves of VAE, $\beta$-VAE, ControlVAE and eVAE on dSprites.}
    \label{fig:eVAE-RD}
\end{figure}

We further compare the VIB effects of eVAE against VAE, $\beta$-VAE and ControlVAE by the rate-distortion (R-D) curve in Figure \ref{fig:eVAE-RD}. VAE (yellow) tends to sacrifice empirical error minimization for representation learning in the early stage, where the rate is optimized to decrease in a quarter of the iterations. As for $\beta$-VAE (green), due to the experiment-specific lower bound ($-\beta_{C} I -D$), the distortion cannot be minimized. ControlVAE (pink) concentrates on optimizing the distortion initially to fit data and optimizing the rate eventually to acquire smooth representation, resorting to the KL-specific lower bound ($-\beta_{KL}I - D$). However, directly modifying $\beta$ by the given PID controller leads to fluctuations between inference capacity and reconstruction quality during the iterations. Rather than monitoring the optimization process, eVAE generates an iteration-specific lower bound ($-\mathcal{E}(\beta_{t})I_{t} -\mathcal{E}(\beta_{t}) D_{t}$) to achieve a balance between minimizing distortion and controlling rate.

\section{Experiments}

We evaluate eVAE in three typical tasks: disentangled representation, image generation, and language  modeling.

\subsection{Dataset and Baselines}
Three different learning tasks and their datasets are listed below to evaluate eVAE against the VAE baselines.

\textbf{Disentangled representation}  learns independent latent variables to generate an image. We validate eVAE on dSprites, a 2D-shape dataset with 737,280 binary images generated by six ground-truth factors: shape (square, ellipse, heart), scale, orientation, position $x$, and position $y$. To examine the disentanglement performance and image generation equally and comprehensively, $\beta$-VAE and ControlVAE are compared.

\textbf{Image generation} reconstructs an imagery sample based on given datapoints. The CelebA dataset (cropped version) \cite{liu2015deep} is used, which is a real-world celebrity face dataset with 202,599 RGB images. The baselines are $\beta$-VAE and ControlVAE.

\textbf{Language modeling} conducts the word-level text generation. The Penn Treebank dataset (PTB) with English corpus \cite{marcinkiewicz1994building} is used. To reveal the performance of restraining KL vanishing, we compare eVAE with three heuristic methods: cost annealing (noted as Cost-10k) \cite{bowman2015generating}, cyclical annealing (Cyc-8) \cite{fu2019cyclical}, and PID control (PID-3) \cite{shao2020controlvae}.

For a fair comparison, we adopt the same embedding network as $\beta$-VAE and tune eVAE to achieve the same KL set point as ControlVAE. The VAE architectures and hyperparameter settings are given in Supplementary Material D.

\subsection{Performance Evaluation}

\begin{figure*}[!htbp]
    \centering
    \includegraphics[width=2.1\columnwidth]{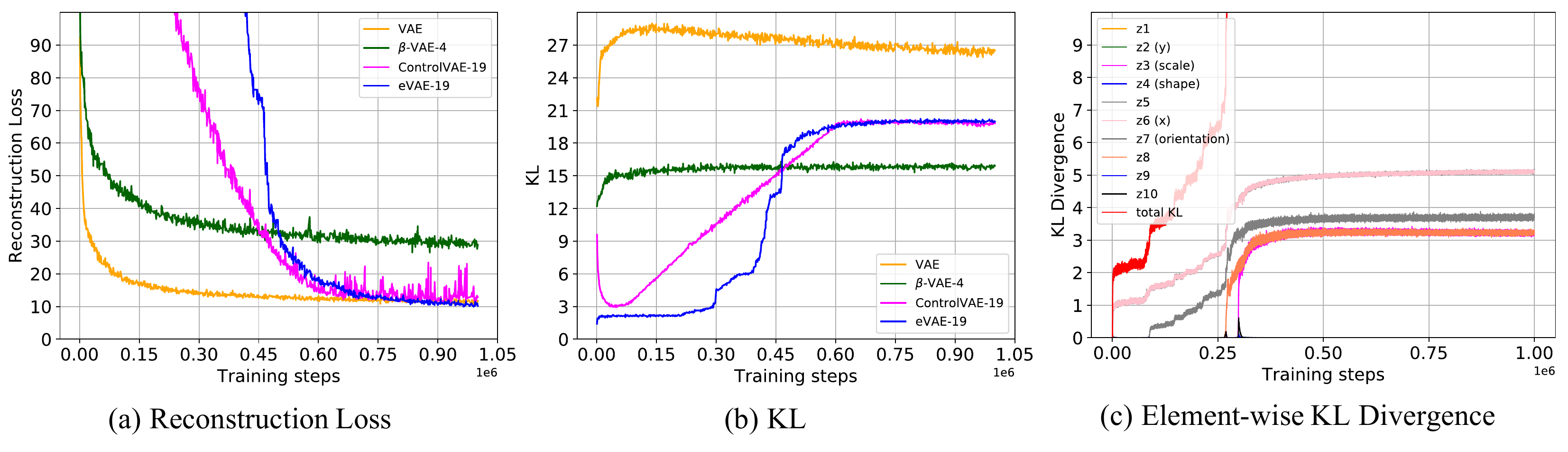}
    \caption{Learning curves on dSprites\footnote{dSprites: Disentanglement testing Sprites dataset. https://github.com/deepmind/dsprites-dataset/}. (a,b) indicate that eVAE has the lowest reconstruction loss compared with VAE ($\beta = 1$), $\beta$-VAE ($\beta = 4$), and ControlVAE (KL = 19) under a fixed KL point KL=19. (c) is the element-wise KL divergence as a function of iterations in eVAE. We can observe that eVAE retains a stable and reasonable KL divergence of each generator dimension (factor): position-$y$ (z2), scale (z3), shape (z4), position-$x$ (z6), orientation (z7). More comparisons in terms of generated KL divergence can be found in Supplementary Material E.More comparisons in terms of generated KL divergence can be found in Supplementary Material E.}
    \label{fig:eVAE-spritCurve}
\end{figure*}

\begin{figure*}[!htbp]
    \centering
    \includegraphics[width=2.1\columnwidth]{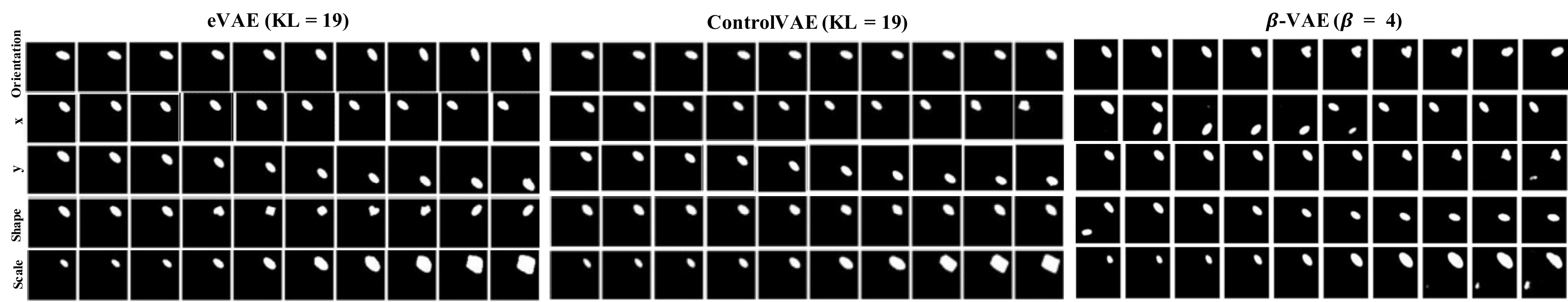}
    \caption{Latent traversal on dSprites. To distinguish the latent factors visually, we take the ellipse for illustration. Each row represents a latent factor in the traversal, while keeping others fixed. The first column refers to the seed image for initialization, and we then manipulate the latent dimension $z$ across the range [-3, 3].
    }
    \label{fig:eVAE-spritfig}
\end{figure*}

\textbf{Performance of disentangled representation learning}
Here, we validate the performance of reconstructing a sharp image with disentangled features. Figure \ref{fig:eVAE-spritCurve}(a) shows that eVAE achieves a lower reconstruction loss of 9.2, compared with ControlVAE under the same expected KL divergence (KL = 19). Moreover, Figure \ref{fig:eVAE-spritCurve}(a) and (b) show that eVAE presents a more stable training curve than other VAEs because of its dynamic weighting guided by the VGA fitness function, rather than a direct modification in $\beta$-VAE and ControlVAE. The variation of each disentangled factor is illustrated in Figure \ref{fig:eVAE-spritCurve}(c). 
Figure \ref{fig:eVAE-spritfig} further demonstrates the reliable disentanglement of all factors and sharp reconstruction quality of eVAE under a set point KL = 19. In contrast, ControlVAE captures the generative factors of scale, $x$ and $y$ while $\beta$-VAE entangles other four factors except scale.

\begin{figure}[!thp]
    \centering
    \includegraphics[width=1\columnwidth]{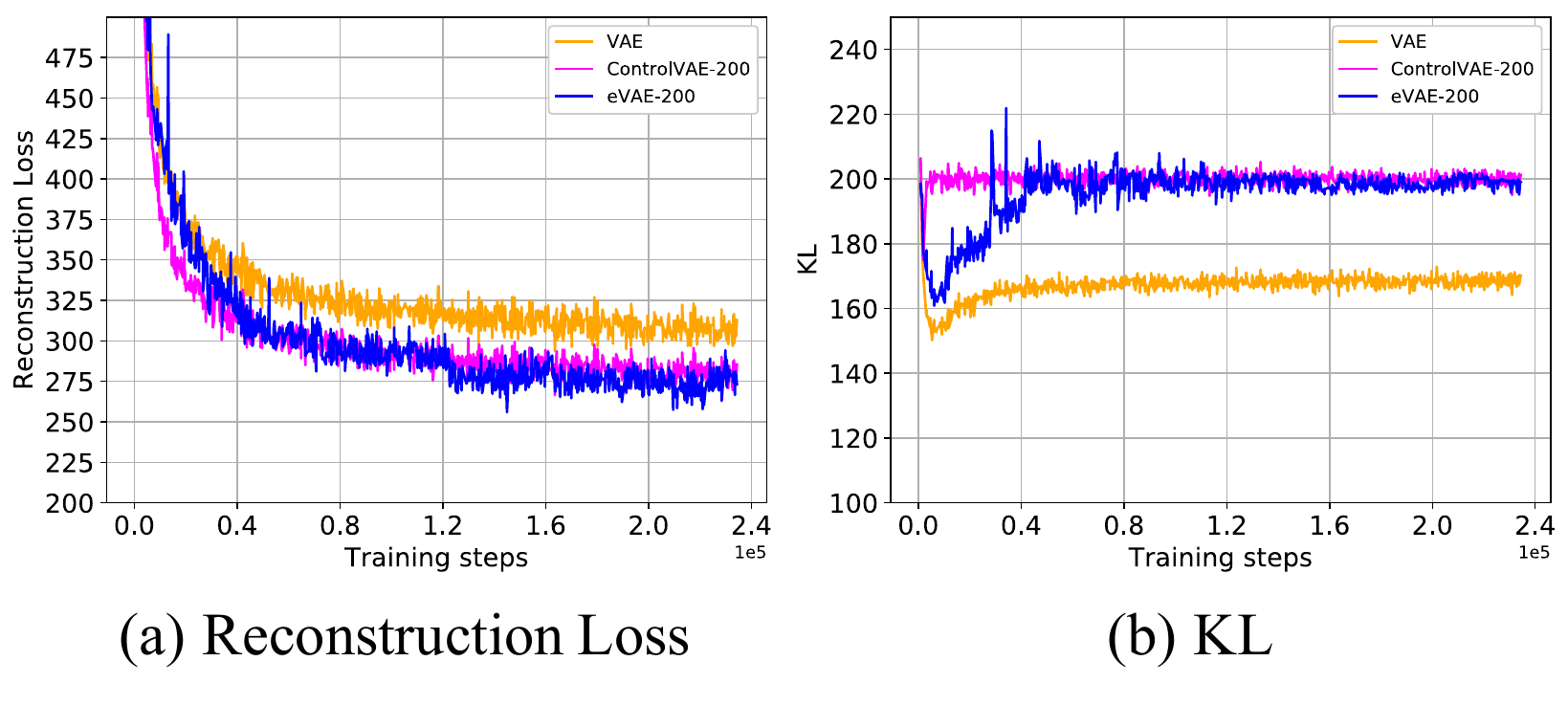}
    \caption{Performance comparison of different VAEs for image generation on CelebA.}
    \label{fig:eVAE-CeleCurve}
\end{figure}

\textbf{Performance of image generation}
We evaluate eVAE on CelebA for image generation. For a fair and concise comparison, the KL set point is set to 200 achieving the best reconstruction quality in \cite{shao2020controlvae}. eVAE is tuned to achieve the desired KL point similar to ControlVAE. Figure \ref{fig:eVAE-CeleCurve}(a) shows that eVAE has the lowest reconstruction error compared with vanilla VAE and ControlVAE. Guided by the  evolving $\beta$, the KL divergence in eVAE can achieve a desired point dynamically, as illustrated in Figure \ref{fig:eVAE-CeleCurve}(b). 

\begin{figure*}[!htbp]
    \centering
    \includegraphics[width=2.1\columnwidth]{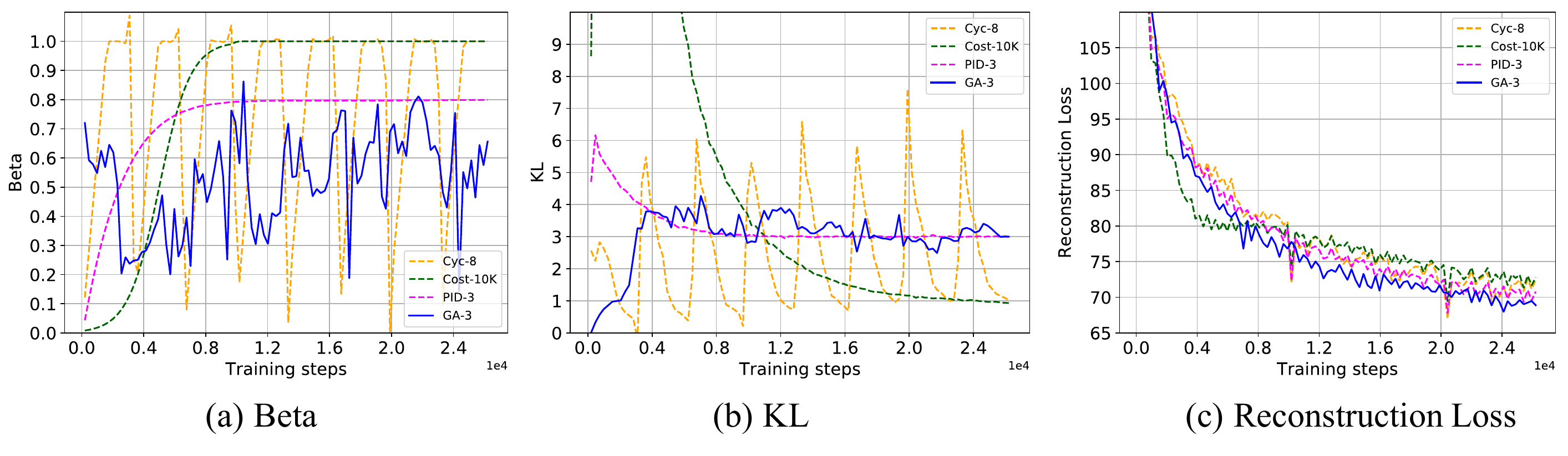}
    \caption{Learning curves on PTB for language modeling. The suffix of PID and VGA refer to a KL set point i.e., KL = 3; Cyc-8 indicates 8 iteration cycles where the weight increases linearly from 0 to 1; Cost-10k refers to 10k iterations during which the weight increases from 0 to 1 by a sigmoid function iteratively.}
    \label{fig:eVAE-nlp}
\end{figure*}

\textbf{Performance of language modeling}
To show the validity of eVAE in preventing the variational language model from degenerating to a standard language model, we draw the learning curves of KL, reconstruction loss, and weight of KL over iterations.
The evolution of $\beta$ is shown in Figure \ref{fig:eVAE-nlp}(a), demonstrating that eVAE can learn to tune $\beta$ over iterations automatically rather than by a heuristic setting. Figure \ref{fig:eVAE-nlp}(b) shows that the Cost annealing (Cost-10k) model still suffers from KL vanishing, with KL approaching 0 at the end of iterations. By contrast, the Cyclical annealing (Cyc-8), PID, and GA can converge to a nonzero divergence, while eVAE achieves the lowest reconstruction loss to 70 in word generation, illustrated in Figure \ref{fig:eVAE-nlp}(c). Further, compared to the PID control, VGA can find a path toward a more stable stage, without undergoing reconstruction fluctuations during training even if the non-zero learned $\beta$ is fixed.

\section{Conclusion}
We propose evolutionary VAE (eVAE) to dynamically learn the tradeoff between reconstruction effectiveness and inference robustness in VAEs. eVAE involves variational evolving operators to solve the premature convergence and random search problem. SGD optimization and genetic algorithm synergistically follow an inner-outer-joint training mechanism in eVAE. Following the variational information bottleneck theory, eVAE derives an iteration-specific lower bound balancing the capacity of compression and decompression over iterations. Multiple task experiments show that eVAE outperforms SOTA VAEs in traversing samples with more disentangled features, generating sharp images with lower reconstruction loss, and resolving the KL vanishing automatically in variational language modeling. We will extend eVAE to other VAE models and tasks. 

\bibliography{evae}
 
\newpage
\clearpage

\section{Supplementary Materials - \\eVAE: Evolutionary Variational Autoencoder}

Here, we present some supplemental materials to the main content, including the eVAE algorithm, the experimental settings, the latent traversal on square and heart, and derivation of the variational lower and upper bounds.

\subsection{A \quad The eVAE algorithm}

Below, Algorithm \ref{algorithm:eVAE} illustrates the modeling process of eVAE, which enables an inner-outer-joint training of VAE and evolutionary learning iteratively.

\begin{algorithm}[H]
    \caption{Evolving Inner-outer-joint Training}
    \label{algorithm:eVAE}
    \renewcommand{\algorithmiccomment}[1]{#1}
    \begin{algorithmic}[1]\label{alg:mt}    
    \STATE \textbf{Input:} crossover rate $Pr_{c}$, mutation rate  $Pr_{m}$,  batch data $\mathbf{x}$. \\
    \STATE \textbf{Output:} $\{\beta^{*},f^{*}\}$ \\
    \STATE \textbf{Initialization:}  parameters of decoder $\theta$, parameters of encoder $\phi$, $\{\beta_l\} = \{\beta_1, \dots, \beta_L\}$ \texttt{//initialize group} \\
      \WHILE{$t < T$}
      \STATE  $Pr_{t}$  $\leftarrow$ sample from $N\left(0, 1\right)$ \texttt{//probability to evolve} \\
      \IF{ $Pr_{t} < Pr_{m}$}
        \STATE $\theta_{t}$, $\phi_{t}$ \texttt{//save current variational parameters} 
        \STATE generate $\beta_{t+1}$ by V-crossover operation $\mathcal{C}(\beta_{t-1},\beta_{t})$ by Eqs. \eqref{con:C1}, \eqref{con:C2} and \eqref{con:C3}
        \STATE evaluate $\beta_{t+1}$ by $f(\mathcal{E}(\phi_t, \theta_t, \{\beta\}, \mathcal{L}_{ELBO_t}))$ by Eqs. \eqref{eq:evae-fitness} and \eqref{con:F2} 
      \ELSIF{$Pr_{t} < Pr_{c}$}
        \STATE $\theta_{t}$, $\phi_{t}$ \texttt{//save current variational parameters} 
        \STATE generate $\beta_{t+1}$ by V-mutation operation $\mathcal{M}(\beta_{t})$ by Eqs. \eqref{con:M1} and \eqref{con:M2}
        \STATE evaluate $\beta_{t+1}$ by fitness function $f(\mathcal{E}(\phi_t, \theta_t, \{\beta\}, \mathcal{L}_{ELBO_t}))$ by Eqs. \eqref{eq:evae-fitness} and \eqref{con:F2}
      \ELSE
        \STATE $\{\beta^{*},f^{*}\} \leftarrow \beta_{l}$ \texttt{//select the strongest pair}
        \STATE $\theta_{t+1}, \phi_{t+1} \leftarrow \mathcal{L}_{eVAE}(\theta_{t},\phi_{t},\beta^{*}|\mathbf{x}_{t})$ \texttt{// update variational parameters}
      \ENDIF
      \STATE \textbf{Return:} $\{\beta^{*},f^{*}\}$ 
      \ENDWHILE
    \end{algorithmic}   
  \end{algorithm}

\subsection{B \quad Derivation of $\beta$-VAE lower bound }
According to the VIB theory, elbo  can be bounded by distortion 
$D=-E_{q_{\phi}(x, z)}[\log p_{\theta}(x \mid z)]$ and rate $R=E_{q_{\phi}(x, z)}[D_{K L}(q_{\phi}(z|x)||p(z))]$, where $I(X,Z)$ has the upper bound $I \leq R$:
\begin{equation}
    \begin{aligned}
    R &= \int q(x) \int q_{\theta}(z \mid x) \log \frac{q_{\theta}(z \mid x)}{p(z)} \\
    &=I(X, Z)+TC(Z) \\
    & \geq I(X, Z)
    \end{aligned}
\end{equation}
and the lower bound $I \geq H -D $:
\begin{equation}
    \begin{aligned}
    I &=\int d x d z p(x, z) \log \frac{p(x \mid z)}{p(x)} \\
    & \geq \int d x d z p(x, z) \log \frac{q(x \mid z)}{p(x)} \\
    &= \int d x d z p(x, z) \log q(x \mid z) \\
    & - \int d x p(x) \log p(x) \\
    & = \int d x d z p(x, z) \log q(x \mid z)\\
    & + H(X) \\
    & = H-D
    \end{aligned}
\end{equation}
All in all, the variational information bottleneck can be regarded as:
\begin{equation}
    H - D \leq I \leq R
\end{equation}
where $H$ is the entropy of the data.
\par 
In $\beta$-VAE framework, $\beta_{C}$ impose a experiment-specific constraint on compressor and decompressor, giving the variational lower bound:
\begin{equation}
    \begin{aligned}
        H - D & \leq I \leq R \\
        H - D & \leq \beta I \leq \beta R \\
        H  & \leq \beta I + D \leq \beta R + D \\
    \end{aligned}
\end{equation}
In summary, the lower bound can be denoted as:
\begin{equation}
    \mathcal{L}_{\beta-VAE}(\theta, \phi)=-D-\beta_{C} R \leq -\beta_{C} I  -D
\end{equation}

\subsection{C \quad Derivation of eVAE lower bound }
According to Eq. \eqref{evae:loss} and Eq. \eqref{eq:evae-fitness} , we can obtain the loss function of eVAE in iteration $t$ as follows:
\begin{equation}
    \begin{aligned}
    \mathcal{L}_{eVAE}  &= E_{\mathbf{z} \sim q_{\phi}(\mathbf{z}_{t} \mid \mathbf{x}_{t})} \log p_{\theta}(\mathbf{x}_{t} \mid \mathbf{z}_{t})+ \\
    & f_{\beta_{t} \sim R}(\beta_{t}, \mathcal{E}) D_{K L}(q_{\phi}(\mathbf{z}_{t} \mid \mathbf{x}_{t}) \| p(\mathbf{z}))  \\
    &= E_{x_{t+1} \sim p_{\mathcal{X}}}[\mathcal{L}_{R} + f(\beta_{t}) \mathcal{L}_{I} + \mathcal{E}(\Delta \mathcal{L}_{E L B O})] \\
    &= E_{x_{t+1} \sim p_{\mathcal{X}}}[\mathcal{L}_{R}+f(\beta_{t}) \mathcal{L}_{I}  \\
    & + \mathcal{E}(\Delta \mathcal{L}_{ELBO_{t+1}} + ||KL_{t+1}(\beta_{t+1}) - c||) ] \\
    \end{aligned}
\end{equation}

According to the VIB theory,  $R_{t}$ sets the upper bound on information bottleneck $I_{t}(X_{t}, Z_{t})$ in $t$-th iteration :
\begin{equation}
    \begin{aligned}
    R_{t} &= \int q(x_{t}) \int q_{\theta}(z_{t} \mid x_{t}) \log \frac{q_{\theta}(z_{t} \mid x_{t})}{p(z_{t})} \\
    &=I_{t}(X_{t}, Z_{t})+TC(Z_{t}) \\
    & \geq I_{t}(X_{t}, Z_{t})
    \end{aligned}
\end{equation}
and $D_{t}$ sets the lower bound:
\begin{equation}
    \begin{aligned}
    I_{t} &=\int d x d z p(x_{t}, z_{t}) \log \frac{p(x_{t} \mid z_{t})}{p(x_{t})} \\
    & \geq \int d x d z p(x_{t}, z_{t}) \log \frac{q(x_{t} \mid z_{t})}{p(x_{t})} \\
    & = \int d x d z p(x_{t}, z_{t}) \log q(x_{t} \mid z_{t}) \\
    & - \int d x p(x_{t}) \log p(x_{t}) \\
    & = \int d x d z p(x_{t}, z_{t}) \log q(x_{t} \mid z_{t})\\
    & + H(X) \\
    & = H - D_{t}
    \end{aligned}
\end{equation}
After optimized by a variational evolutionary learner $\mathcal{E}$, the loss function of eVAE constrained by a dynamic lower bound is as follows: 
\begin{equation}
    \begin{aligned}
    \mathcal{L}_{e V A E}  &= E_{\mathbf{z} \sim q_{\phi}\left(\mathbf{z}_{t} \mid \mathbf{x}_{t}\right)} \log p_{\theta}(\mathbf{x}_{t} \mid \mathbf{z}_{t})+ \\
    & f_{\beta_{t} \sim R}\left(\beta_{t}, \mathcal{E}\right) D_{K L}(q_{\phi}(\mathbf{z}_{t} \mid \mathbf{x}_{t}) \| p(\mathbf{z})) \\
    &= E_{x_{t+1} \sim p_{\mathcal{X}}}[\mathcal{L}_{R} + f(\beta_{t}) \mathcal{L}_{I} + \mathcal{E}(\Delta \mathcal{L}_{E L B O})] \\
    &= E_{x_{t+1} \sim p_{\mathcal{X}}}[\mathcal{L}_{R}+f(\beta_{t}) \mathcal{L}_{I}  \\
    & + \mathcal{E}(\Delta \mathcal{L}_{ELBO_{t+1}} + ||KL_{t+1}(\beta_{t+1}) - c||) ] \\
    &= -D_{t} - \beta_{t}R_{t} -\mathcal{E} (-D_{t+1} - \beta_{t+1}R_{t+1} + D_{t}+ c) \\
    & \leq -\mathcal{E}(\beta_{t}) D_{t} - \mathcal{E}(\beta_{t}) I_{t}\\
    \end{aligned}
\end{equation}
where $c$ is the desired KL point and $\beta_{t} $ is a chromosome sample in $t$-th iteration.

\subsection{D \quad Experimental settings}

Table \ref{tab:exp} shows the experimental settings of the three learning tasks for evaluating eVAE. We give the settings of the encoder-decoder network architectures and the parameters for the evolutionary learning on three datasets dSprites, CelebA, and PTB and their corresponding optimizers. 
\begin{table*}[htbp]
\resizebox{0.95\textwidth}{!} 
{ 
\begin{tabular}{@{}llllll@{}}
\toprule
\textbf{Dataset} & \textbf{Optimiser} &  & \textbf{Architecture} & \textbf{eVAE initialization} &  \\ \midrule
dSprites &
  Adam(1e-4) &
  \begin{tabular}[c]{@{}c@{}}Input\\ Encoder\\ \\ \\ \\ Latent\\ Decoder\end{tabular} &
  \begin{tabular}[c]{@{}l@{}}(64,64,1)\\ Conv(32, 32, 32), ReLu, Conv(32,16,16), \\ ReLu, Conv(32,8,8), ReLu,Conv(32,4,4), \\ Relu, Linear(256), Relu, Linear(256),\\  Relu, Linear(20)\\ 10\\ Linear(256), Relu, Linear(256),\\  Relu, Linear(512), Relu, \\ Conv(32,4,4), Relu, Conv(32,8,8),\\  Relu, Conv(32,16,16), Relu,\\  Conv(32,32,32), Relu, (1, 64,64)\end{tabular} &
  \begin{tabular}[c]{@{}l@{}}Mutation Rate\\ Crossover Rate\\ Individual Number\end{tabular} &
  
  \begin{tabular}[c]{@{}l@{}}0.001\\ 0.04\\ 20\end{tabular} \\
  \hline 
\begin{tabular}[c]{@{}l@{}}CelebA\\ (KL = 200)\end{tabular} &
  Adam(1e-4) &
  \begin{tabular}[c]{@{}l@{}}Input\\ Encoder\\ \\ \\ \\ \\ Latent\\ Decoder\end{tabular} &
  \begin{tabular}[c]{@{}l@{}}(128,128,3)\\ Conv(32, 32, 32), ReLu,\\  Conv(32,16,16), ReLu,\\ Conv(32,8,8),ReLu,\\ Conv(32,4,4), Relu, Linear(256),\\ Relu, Linear(256), Relu, Linear(20)\\ 10\\ Linear(256), Relu, Linear(256),\\ Relu, Linear(512), Relu, \\ Conv(32,4,4), Relu, \\ Conv(32,8,8), Relu, \\ Conv(32,16,16), Relu, Conv(32,32,32), \\ Relu, (3, 64,64)\end{tabular} &
  \begin{tabular}[c]{@{}l@{}}Mutation Rate\\ Crossover Rate\\ Individual Number\end{tabular} &
  \begin{tabular}[c]{@{}l@{}}0.001\\ 0.03\\ 20\end{tabular} \\
   \\
   \hline 
\multicolumn{1}{l}{\begin{tabular}[c]{@{}l@{}}PTB\\ (KL = 3)\end{tabular}} &
  \multicolumn{1}{l}{\begin{tabular}[c]{@{}l@{}}init\_lr: 0.001\\ threshold:100\\ decay\_factor: 0.5\\ max\_decay: 5\end{tabular}} &
  \multicolumn{1}{l}{\begin{tabular}[c]{@{}l@{}}Input\\ Encoder\\ Latent\\ Decoder\end{tabular}} &
  \multicolumn{1}{l}{} &
  \multicolumn{1}{l}{\begin{tabular}[c]{@{}l@{}}Mutation Rate\\ Crossover Rate\\ Individual Number\end{tabular}} &
  \multicolumn{1}{l}{\begin{tabular}[c]{@{}l@{}}0.006\\ 0.07\\ 10\end{tabular}} \\ \bottomrule
\end{tabular}
}
\caption{Experimental settings for disentangled representation learning on  dSprites, image generation on CelebA, and language generation on PTB.}
\label{tab:exp}
\end{table*}

\subsection{E \quad Analysis of element-wise KL divergence in disentangled representation learning}

\begin{figure*}[!htbp]
    \centering
    \includegraphics[width=2\columnwidth]{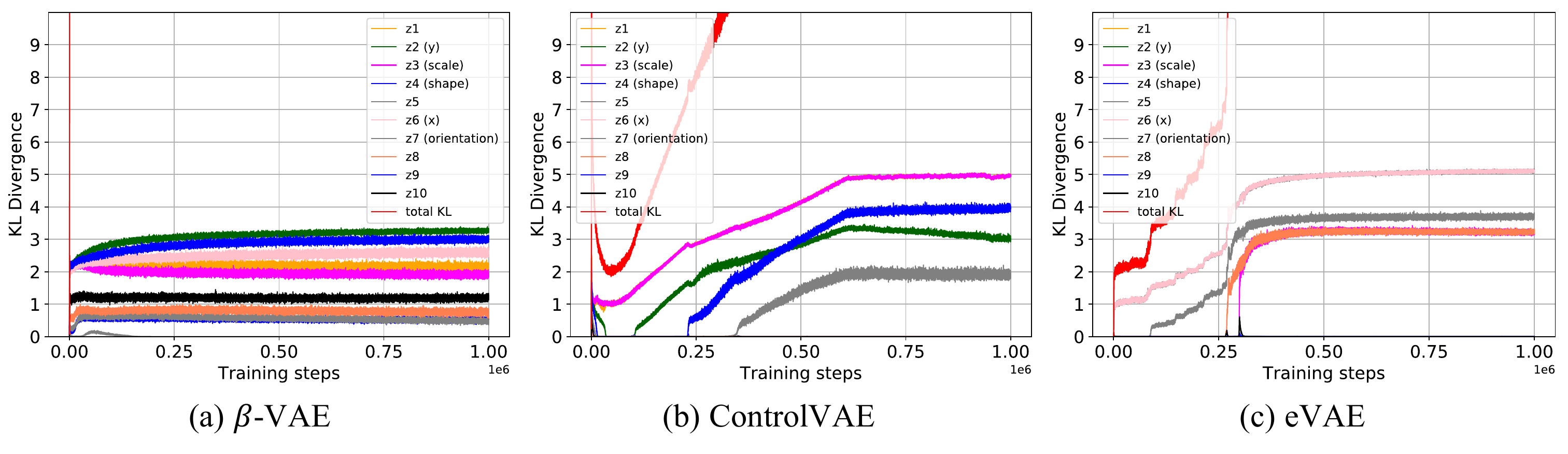}
    \caption{The element-wise KL divergence over iterations for $\beta$-VAE, ControlAVE, and eVAE on dSprites. It shows that each KL divergence of eVAE is larger than that of $\beta$-VAE, providing an enough space for generating disentangled representation. In addition, eVAE generates more stable KL divergence compared to ControlVAE.}
    \label{fig:eVAE-kl}
\end{figure*}

Figure \ref{fig:eVAE-kl} further shows the element-wise KL divergence evolving over iterations of three VAE models: $\beta$-VAE and ControlAVE in comparison with our eVAE on the dataset dSprites.

\end{document}